\pdfoutput=1
\documentclass[11pt]{article}
\usepackage[final]{acl}

\usepackage{times}
\usepackage{latexsym}
\usepackage[T1]{fontenc}
\usepackage[utf8]{inputenc}
\usepackage{microtype}
\usepackage{inconsolata}

\usepackage{graphicx}

\usepackage{hyperref}
\usepackage{url}
\usepackage{tikz}
\usepackage{booktabs}
\usepackage{amsmath,amssymb}
\usepackage{svg}
\usepackage{makecell}
\usepackage{multirow}
\svgpath{{figures/}} 
\usepackage[textsize=tiny]{todonotes}

\title{Exploring Diachronic and Diatopic Changes in Dialect Continua: \\Tasks, Datasets and Challenges}

\author{Melis Çelikkol\thanks{These authors contributed equally to this work.}$^{1}$ \, 
    Lydia K\"orber\footnotemark[1]$^{1}$ \, 
    Wei Zhao$^{2}$ \\[0.4em]
    $^{1}$University of Heidelberg\,  
    $^2$University of Aberdeen \\
    \texttt{\{firstname.lastname\}@stud.uni-heidelberg.de}\\
    \texttt{wei.zhao@abdn.ac.uk}\\
  }

\begin{document}
\maketitle
\begin{abstract}
Everlasting contact between language communities leads to constant changes in languages over time, and gives rise to language varieties and dialects. However, the communities speaking non-standard language are often overlooked by non-inclusive NLP technologies. Recently, there has been a surge of interest in studying diatopic and diachronic changes in dialect NLP, but there is currently no research exploring the intersection of both. Our work aims to fill this gap by systematically reviewing diachronic and diatopic papers from a unified perspective.
In this work, we critically assess nine tasks and datasets across five dialects from three language families (Slavic, Romance, and Germanic) in both spoken and written modalities. The tasks covered are diverse, including corpus construction, dialect distance estimation, and dialect geolocation prediction, among others. Moreover, we outline five open challenges regarding changes in dialect use over time, the reliability of dialect datasets, the importance of speaker characteristics, limited coverage of dialects, and ethical considerations in data collection. We hope that our work sheds light on future research towards inclusive computational methods and datasets for language varieties and dialects.

\end{abstract}

\section{Introduction}
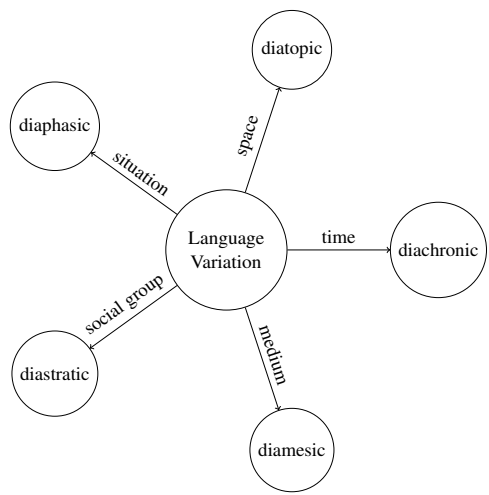
\begin{figure}[htbp]
    \centering
    \resizebox{0.4\textwidth}{!}{
    \begin{tikzpicture}
  \node[circle, draw, text width=2cm, align=center] (center) at (0,0) {Language Variation};
  \foreach \angle/\nodelabel/\edgelabel in {0/diachronic/time, 72/diatopic/space, 144/diaphasic/situation, 216/diastratic/social group, 288/diamesic/medium} {
    \node[circle, draw] (node\nodelabel) at (\angle:4.3) {\nodelabel};
    \draw[->] (center) edge node[midway, sloped, above] {\edgelabel} (node\nodelabel);
  }
    \end{tikzpicture}
    }
    \caption{Language variation and the diasystem.\protect\footnotemark}

    \label{fig:diagram}
\end{figure}

Language continuously changes, varies and transforms on all levels of linguistics. Research in sociolinguistics assumes five dimensions of language variation, the so-called diasystem, that are mutually influential: diaphasic (situation), diamesic (medium), diastratic (social group), diachronic (time), and diatopic (space), as shown in Figure \ref{fig:diagram} \citep{zampieri2020natural}. 

\footnotetext{Inspired by: \url{http://phylonetworks.blogspot.com/2015/06/the-diasystematic-structure-of.html}, accessed on 11.03.2024.}

Diaphasic, diamesic, diastratic and diatopic variation can be grouped to synchronic variation, as opposed to diachronic variation which spans several points in time. Diachronic variation is not limited to decades and centuries, but may already be observed within years, months, and even weeks or days.
Especially with computer-mediated communication and social media platforms, language change appears to spread at a faster pace \cite{10.1371/journal.pone.0113114}. 
This exposes a challenge in NLP applications, as models remain static after training and struggle to understand 
the evolving nature of language\footnote{
Although there are methods to keep models up-to-date, such as re-training, fine-tuning, and RAG (Retrieval-Augmented Generation) leveraging up-to-date information sources at inference, the process is time-consuming and costly.}. As a result, model performance decreases over time.
For instance, headline generation models decrease in performance after a few years, while emoji prediction models do so even within a month \cite{sogaard-etal-2021-need}.
As shown in the (socio-)linguistic work \citep{beeching2006synchronic},
diachronic and synchronic variation are closely linked in the sense that
language change often manifests first in synchronic variation before entering a diachronic level. Additionally, there is a strong spatial component in language change, as 
language change
is caused by contact between people and speech communities (lately by online interactions too), which gives rise to dialects \cite{Jeszenszky_Stoeckle_Glaser_Weibel_2018}. 
While \textit{isoglosses} separate dialects by drawing the geographic boundaries, 
the consensus among dialectologists and sociolinguists today is to speak of \textit{dialect continua}, which assume gradual transitions between central areas of different dialects over time
\cite{Jeszenszky_Stoeckle_Glaser_Weibel_2018}. 
In these continua, 
as proposed by the \textit{wave model} \cite{wolfram_schillingch24}, language change is propagated from a certain locus at a certain point in time and spread layer-wise, radiating from the central point of contact. This is indeed a result of both spatial (diatopic) and temporal (diachronic) interactions within dialect continua.

\begin{figure}
    \centering
    \includegraphics[width=0.48\textwidth]{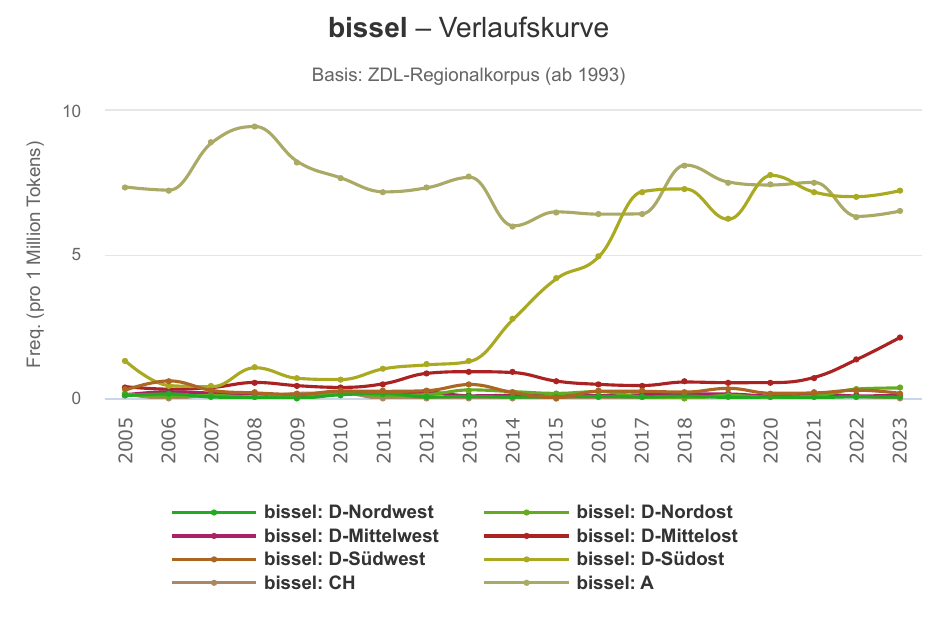}
    \caption{Diachronic usage of \textit{bissel} in the years 2005-2023 in a regional newspaper corpus of German across dialect areas in frequency per 1M tokens.\protect\footnotemark}
    \label{fig:bissel}
\end{figure}

An example of diatopic variation over time can be seen in Figure \ref{fig:bissel}: the usage of the German dialect word \textit{bissel} (a bit). A query in the ZDL-Regionalkorpus \citep{NoldaBarbaresiGeyken2021, NoldaBarbaresiGeyken2023}, a collection of regional newspaper texts from Germany, Austria, and Switzerland, reveals its constant usage in Austria (A), and an increased usage in other, more northern regions over time, first in Bavaria (D-Südost) possibly due to geographic proximity, and a more recent rise in Central Eastern Germany (D-Mittelost).

In this work, we explore the intersection of diachronic and diatopic changes in language variants and dialects within the NLP community. To do so, we investigate nine tasks and datasets across five dialects from three language families to address the following research questions:

\begin{itemize}
    \item What are the characteristics of dialect datasets across different time periods and geographic areas, and what NLP tasks have been established based on these datasets (\S\ref{sec:methods})?
    \item What is the current state of computational methods and their results in these dialect-related NLP tasks (\S\ref{sec:results})?
    \item What are the challenges in dialect NLP research that have not been addressed in previous works (\S\ref{sec:challenges})?
\end{itemize}

\footnotetext{Usage graph for \textit{bissel}, created with Digitales W\"orterbuch der deutschen Sprache (DWDS, Digital Dictionary of the German Language), \url{https://shorturl.at/9XVwt}, accessed on 04.07.2024.}

In this work, we aim at exploring the intersection of diachronic and diatopic variation in dialect NLP research. 
Research questions on this topic include (a) how to detect and quantify language change in dialect continua over time,
and (b) how to build and process diachronic-diatopic datasets.
Previous approaches leveraged machine learning methods to compute the distance and similarity of different varieties on various linguistic levels such as graphemics \citep{WaldenbergerDipperLemke+2021+401+420}, syntax \citep{Jeszenszky_Stoeckle_Glaser_Weibel_2018, chen2024syntactic}, phonetics
\citep{boldsen-paggio-2022-letters, he2024sound}, semantics \citep{montanelli2023survey, ma-etal-2024-graph, ma2024presence}, and built diachronic-diatopic datasets in both written and spoken modalities \citep{koprivova-etal-2014-mapping, komrskova}.

Here, we critically review nine tasks and datasets, highlighting their strengths and limitations, as well as identifying challenges that have not been previously addressed.
We discuss seminal works in Indo-European languages and their varieties, as well as recent works on this topic. 
The dialect continua covered here include the Slavic family with the Czech dialect landscape \cite{koprivova-etal-2014-mapping, komrskova},  the Romance language family with Italian \cite{ramponi-casula-2023-diatopit} and Portuguese \cite{pichel-campos-etal-2018-measuring, zampieri-etal-2016-modeling}, as well as the Germanic language family with Swiss German \cite{Jeszenszky_Stoeckle_Glaser_Weibel_2018, jeszenszky-etal-2019-spatio} and historical German varieties \cite{dipper-waldenberger-2017-investigating, WaldenbergerDipperLemke+2021+401+420}. 

\section{Related work}
To our knowledge, there is no survey examining the intersection of diachronic and diatopic variation in dialect NLP so far. However, there are survey papers examining the diachronic and diatopic components separately, which will be briefly presented here.
Diachronic language modeling has been surveyed with regard to embeddings \citep{kutuzov-etal-2018-diachronic} and semantic shift detection \citep{montanelli2023survey}. 

The comprehensive survey on diatopic language modelling by \citet{zampieri2020natural} evaluates computational methods for processing similar languages, language varieties, and dialects, with a  focus on diatopic language variation and integration in NLP applications. The work identifies the availability of suitable data as a key challenge, as the classical NLP data sources like newspaper text and Wikipedia do not cover dialectal data. Instead, social media posts and speech transcripts can be used. More recently, an evaluation benchmark for different NLP tasks in dialects, varieties and closely-related languages, \textsc{DialectBench}, was published \cite{faisal2024dialectbench}, proving that variation is of current interest in the research community. \citet{joshi2024natural} survey Natural Language Understanding and Generation in dialects, without taking other axes of the variation diasystem into account.
There exists a designated series of workshops on NLP for Similar Languages, Language Varieties, and Dialects (VarDial)\footnote{cf. 2024 edition \url{https://sites.google.com/view/vardial-2024/home}, accessed on 11.03.2024.}, which also proposes several NLP shared tasks 
in dialects and other varieties, 
such as dialect classification and identification itself. Even though the workshop has featured a number of publications and talks dealing with the intersection of diachronic and diatopic variation over the years \cite{sukhareva-chiarcos-2014-diachronic, baldwin-2018-language, vidal-gorene-etal-2020-recycling}, this has not been a separate workshop or shared task topic up until now.

\section{Tasks and Datasets}\label{sec:methods}
\begin{table*}[h]
\setlength{\tabcolsep}{3pt}
    \footnotesize
    \centering
    \begin{tabular}{lll} \toprule
         \textbf{Languages} & \textbf{Tasks} & \textbf{Datasets}\\ \midrule
         Czech &  Corpus Construction \cite{koprivova-etal-2014-mapping, komrskova} &  ORTOFON, DIALEKT\\
         \midrule
         Italian  & Geolocation Prediction \cite{ramponi-casula-2023-diatopit} & \textsc{DiatopIt} \\
         \midrule
         Portuguese   & Language Distance Estimation \cite{pichel-campos-etal-2018-measuring} & DiaPT \\
         \midrule
         Portuguese  & Century Classification 
         \cite{zampieri-etal-2016-modeling}& Colonia \\
         \midrule
         Swiss German
          &  Modeling of Dialectal Variant Transition \cite{Jeszenszky_Stoeckle_Glaser_Weibel_2018} & SADS \\
          \midrule
          Swiss German & Predicting Which Regions Use Which Dialectal Variants \cite{jeszenszky-etal-2019-spatio} & SADS \\
         \midrule
         German
         & Investigating Diachronic Changes in Dialects \cite{dipper-waldenberger-2017-investigating} & Anselm \\
         \midrule
         German & Investigating Graphemic Variation in Dialects \cite{WaldenbergerDipperLemke+2021+401+420} & ReM \\
         \midrule
         English, French & Semantic Change Detection \cite{montariol-allauzen-2021-measure} & Le Monde, NY Times\footnotemark \\
         \bottomrule
    \end{tabular}
    \caption{An overview of the presented papers in Section \ref{sec:methods}. 
}
    \label{tab:methods}
\end{table*}

In this section, we review the dialect-related tasks and datasets from a unified perspective considering both diachronic and diatopic aspects, and organize them by languages (See Table \ref{tab:methods} for tasks and datasets, and Table \ref{tab:my_label} for data statistics).

\subsection{Czech}
A very interesting albeit not very recent paper by \citet{koprivova-etal-2014-mapping} explains the building process of their later released ORTOFON and DIALEKT corpora \cite{komrskova}. Although both papers are mention-worthy, we focus on \citet{koprivova-etal-2014-mapping} due to the presentation and depth of explanation for the data collection processes.

The ORTOFON corpus relies on spontaneous conversations recorded between 2012-2017, where nobody was aware that the conversations were recorded except for the person who made the recording.
The non-scripted interactions recorded this way are then separated into the closest one of 12 situation categories which were created with the topics of Czech daily-life in mind. What makes this corpus really strong is that \citet{koprivova-etal-2014-mapping} consider several missing elements 
in other corpora all at once: relationship between speakers is noted alongside the total number of generations present in each conversation, as well as the speaker characteristics, such as education, occupation, region of residence (with subtypes such as longest, childhood and current residence) and speech defects. After factoring these elements into the data collection process,
the corpus is balanced according to the speaker's gender, education (binary as tertiary/non-tertiary), age (binary as $>$35 or $<$35), and childhood region of residence. 
As promising as \citet{koprivova-etal-2014-mapping}'s collection methods are to provide natural results, the approach is not discussed in terms of ethics in their presentation. 

DIALEKT on the other hand presents a collection
of 
regional dialects from the 1960s-1980s. The DIALEKT corpus includes dialects, some of which are even extinct now. The DIALEKT monologues are all by people who have always lived in rural areas and are all natives to their regions. One can argue that DIALEKT also considers generational difference,
as the birth years of speakers range from
the end of 19th century to the start of 20th century, although may not be to the extent of ORTOFON in some cases. Another feature of DIALEKT worth mentioning is that it allows users to search for dialect features captured with regards to all levels of linguistic analysis.  

Both corpora utilize ELAN linguistic transcription software \citep{sloetjes2008annotation}, going through annotation in two tiers. For ORTOFON, the first layer is close to Czech orthography while the second adapts phonetic transcription. The latter enables collecting features such as stress groups, vowel reductions and cliticization which might have been lost otherwise. For DIALEKT, the first layer is dialectological, and the second is the ortographic one same as ORTOFON. In this case, the dialectological layer allows distinguishing speech sounds which are special to non-standard varieties of Czech via the use of a set of symbols. These qualities make the corpora later presented by \citet{komrskova} worth of note.

\subsection{Italian}
Recently, \citet{ramponi-casula-2023-diatopit} present DIATOPIT, a corpus built by analyzing Twitter posts of non-Standard Italian use. 
They use Twitter APIs to locate non-standard use of language across Italian borders. Moreover, they collect data that comes from accurate coordinates throughout two years to ensure no occasional visitors will disturb the data. They also consider a variety of ``out of vocabulary'' (OOV) tokens that they use to deduct which of the Twitter posts collected may be from a regional language user. OOV tokens contain tokens which may not be special tokens (i.e. hashtag) and also may not exist in the Aspel dictionary for Italian, but do not include common interjections, elongated words, slangs, wrong diacritics or foreign language tokens, as well as named entity tokens. In doing so, the coordinates from Twitter API and the OOV tokens can be matched to create a map of data by the administrative region.

They
also include experiments for evaluating the DIATOPIT's representativeness of real varieties of Italian,
which is shown to yield satisfying results in their metrics. While they 
list a variety of goals for their corpus, what we can say truthfully is that the main contribution is to enable a starting point for those interested in applying NLP methods to research varieties of dialects spoken within Italy. It also serves as first example focusing Italian diatopic variation. 
\footnotetext{These corpora are not listed in Table \ref{tab:my_label}, as they are not described in detail.}

\subsection{Portuguese}
A different approach works with historical Portuguese to identify 
different time periods within the historical evolution of a language. \citet{pichel-campos-etal-2018-measuring} use a perplexity based measure for this task. Perplexity is a metric indicating how well a system fits a text sample, with a lower score being the better score. It is commonly used as a measure to evaluate the quality of a system, \citet{pichel-campos-etal-2018-measuring} note that this is the first attempt utilizing perplexity 
to calculate diachronic language distance between different periods of historical Portuguese. Their corpus includes six time periods of European Portuguese ranging from the 12th century to the 20th century. They collect their data from various open historical text repositories and historical corpora, and keep the original spelling whenever possible. 
The perplexity-based approach is noted to successfully identify three main periods for European Portuguese, and should be applicable with other languages as well.

There is another study that works with Portuguese: 
\citet{zampieri-etal-2016-modeling} build upon the Colonia corpus that 
is an already existing historical Portuguese corpus with texts from the 16th century to the early 20th century. Additionally, they include Part-of-Speech tags for the corpus.

\subsection{Swiss German}
An interesting approach of modeling transition areas between different dialectal variants using logistic functions is proposed by \citet{Jeszenszky_Stoeckle_Glaser_Weibel_2018}: The idea is to model geographic areas, where one dialectal variant transitions into another, i.e. where language change is taking place. They base their analyses on the SADS dataset \citep{GlaserBart+2015+81+108}, a linguistic survey with questions on different dialectal phenomena in Swiss German which provides detailed geolocations. 
Even though the method is very elaborate on a geo-linguistic level, a major drawback is that it can only model the transition of two variants, whereas in real-world scenarios, variation patterns are much more complex and numerous variants are assumed to coexist and influence one another.
In a subsequent study on the same dataset, the authors focused further on the temporal aspect \cite{jeszenszky-etal-2019-spatio}, and also took the age of respondents into account, an approach similar to \citet{koprivova-etal-2014-mapping}. With the sociolinguistic diasystem of language variation, these studies model not only two, but three dimensions: diachronic, diatopic, and diastratic by taking the social variable of age into account.

\begin{table*}[h]
    \footnotesize
    \centering
    \begin{tabular}{l l l l l l l l} \toprule
         \textbf{Languages} & \textbf{Datasets} & \textbf{Tokens} & \textbf{Source/Register} & \textbf{Time Span} & \textbf{Modality}\\ \midrule
         Czech & ORTOFON \citep{komrskova} & 1.24 M & dialogue & 2012-2017 & spoken  \\
         Czech & DIALEKT \citep{komrskova} & 126,131 & monologue & 1960s-1980s & spoken \\
         Italian & \textsc{DiatopIt} \citep{ramponi-casula-2023-diatopit} & 388,069 & Twitter & 2020-2022 & written \\
         Portuguese & DiaPT \citep{pichel-campos-etal-2018-measuring} & - & historical text & 1100-2000 & written\\
         Portuguese & Colonia \citep{zampieri2013colonia} & 5.1 M & media, historical text & 1500-2000 & written \\
         Swiss German & SADS \citep{GlaserBart+2015+81+108} & -& linguistic survey & 2000-2002 & written  \\
         German & Anselm \citep{dipper2013anselm} & 30,000 & religious text & 1350-1600 & written \\ 
         German & ReM \citep{Petran_Bollmann_Dipper_Klein_2016} & 2.5 M & historical text & 1050-1350 & written \\
         German & ZDL-Reg. \citep{NoldaBarbaresiGeyken2021} & 11.78 B & regional newspaper & 1993-2024 & written \\
         \bottomrule
    \end{tabular}
    \caption{An overview of the dialect-related datasets discussed in Section \ref{sec:methods}. 
    ZDL-Reg. is dynamically enlarged; the number of tokens is taken from \url{https://www.dwds.de/d/korpora/regional}, accessed on 05.03.2024.
    SADS does not contain natural language data, but 118 multiple-choice questions about 54 (morpho-)syntactic phenomena. Additionally, we include another corpus of regional newspaper data in German, the ZDL-Regionalkorpus \cite{NoldaBarbaresiGeyken2021, NoldaBarbaresiGeyken2023}---which has not been explored for diachronic-diatopic studies yet.}
    \label{tab:my_label}
\end{table*}

\subsection{German}
There are two noteworthy diachronic-diatopic studies on historical corpora of German: 
\citet{dipper-waldenberger-2017-investigating} examine language change across dialects on a graphemic level. They use aligned equivalent word forms (i.e. word forms that have the same normalization to Standard German) from different German regions to derive rewrite rules with insertions, replacements and identity and create mappings based on weighted Levenshtein Distances. The results show differences across
linguistic levels including morphology, phonology and graphemics. 
The results align with findings from historical linguistics on specific phenomena, such as the High German consonant shift.
A follow-up work by \citep{WaldenbergerDipperLemke+2021+401+420}
uses a different dataset, Reference Corpus of Middle High German (Referenzkorpus Mittelhochdeutsch, ReM) \citep{Petran_Bollmann_Dipper_Klein_2016}, and generate difference profiles based on weighted Levenshtein distance. The work includes
word boundaries as well which allows for capturing further linguistic phenomena. The created mappings from one historical and dialectal variety to another are then compared on a graphemic and graphophonemic level. On a broader level, they conduct further statistical analyses by comparing the intersection of shared mappings between texts in a diatopic subcorpus, and find that this measure indeed reflects the similarity of neighboring dialects.

\subsection{English and French}
An example of using diachronic word embeddings to model semantic change in 
the English and French languages is the work by \citet{montariol-allauzen-2021-measure}. Although this work does not work with dialectal data, we still decided to include it, as the approach is interesting and could be applied to (non-continuous) dialect data, e.g. Standard German and Swiss German. Since the datasets are not described in detail, we decided to not include them in Table \ref{tab:my_label}. Overall, the work proposes learning word embeddings from a synthetic corpus with the CBOW (continuous bag-of-words) approach and M-BERT \citep{devlin-etal-2019-bert}, and experiments with different training and aggregation techniques. Computing the divergence of word senses in the two languages, they analyze different language change patterns such as stability in both languages, drift in the same direction, and divergence in word senses with culture-specific contexts.
\citet{cathcart-wandl-2020-search} propose a related approach experimenting with word embeddings to model phonological change in related varieties of historical Slavic languages in a continuous and discrete way. These approaches are quite interesting and could be applicable to dialect data as well, given the availability of a large amount of training data for dialect embeddings and an evaluation corpus that includes sense and phonetic information.

\subsection{Data Characteristics}

\paragraph{Data Sources.} Different text sources have been used for collecting diatopic datasets: While some approaches work with social media data from Twitter \citep{dunn-wong-2022-stability, ramponi-casula-2023-diatopit}, historical corpora mainly contain religious text or official documents \citep{dipper-waldenberger-2017-investigating, WaldenbergerDipperLemke+2021+401+420} and are usually not suited for a geographical analysis on a fine-grained level. The approaches working on Swiss German \citep{Jeszenszky_Stoeckle_Glaser_Weibel_2018, jeszenszky-etal-2019-spatio} do not base their analyses on natural language data, but on a linguistic multiple-choice survey, the Syntactic Atlas of German-speaking Switzerland (SADS). This kind of data can still be very useful, as it provides direct information about specific language phenomena paired with a very fine-grained, reliable geolocation.

\paragraph{Modality.} Most of the corpora rely on written language, only \citet{koprivova-etal-2014-mapping} create two spoken language corpora. From a linguistic point of view, this is very effective, since variation usually is much stronger in spoken compared to written language, as most dialects do not deviate markedly from Standard languages in the written modality.

\paragraph{Time Span.} The diachronic spans of the datasets also vary strongly: While some historical corpora cover very long periods of time, e.g. the Diachronic Portuguese Corpus (DiaPT) \cite{pichel-campos-etal-2018-measuring} spans almost one millennium, social media-based corpora like \textsc{DiatopIt} or linguistic survey data like SADS only span two years.

\paragraph{Data Imbalance.} It must be noted that the Colonia corpus used by \citet{zampieri-etal-2016-modeling} does not contain the same amount of text from each period it covers. For instance,
there are 38 documents available from the 19th century, while there are only 13 available from the 16th century.
Due to this, \citet{zampieri-etal-2016-modeling} generate artificial texts with around 330 tokens for their train and test sets in order to conduct their main experiments.

\section{Experiments}
\label{sec:results}

Experimental setups and results of the presented studies are difficult to compare, as the tasks and datasets presented in Section \ref{sec:methods} are very different. Some of the papers focus on corpus construction \citep{koprivova-etal-2014-mapping} or qualitative analysis \citep{dipper-waldenberger-2017-investigating}, while some 
present quantitative results in the tasks of
measuring language distance, predicting geolocation or dialect variant usage, will briefly be compared here.

\paragraph{Czech.} Since \citet{koprivova-etal-2014-mapping} aims to build/present corpora, there are no experiments to mention. But one can argue that when ORTOFON and DIALEKT are used interconnectedly, they will present a good outlook on diachronic and diatopic variation in Czech. The work by \citet{koprivova-etal-2014-mapping} is to set apart with their detailed annotation system separated with several parallel layers to accommodate speakers individually. In the follow-up work by \citet{komrskova}, the advantages are evident thanks to the use of this multi-tier transcription. 

\paragraph{Italian.} \citet{ramponi-casula-2023-diatopit} evaluate geolocation predictions on two levels: coarse-grained geolocation (CG, i.e. region classification), and fine-grained geolocation (FG, double-regression i.e. for latitude/longitude coordinates). They measure the accuracy of the prediction results in the macro-averaged Precision, Recall, and F1 score.
Baseline models are mostly built upon BERT \citep{devlin-etal-2019-bert}. Both monolingual (Italian-only) and multilingual models are investigated, including AlBERTo \citep{polignano2019alberto}, UmBERTo \citep{musixmatch-2020-umberto} and mBERT \citep{devlin-etal-2019-bert}, XLM-R \citep{conneau-etal-2020-unsupervised}.
Additionally, for CG they use 
Logistic Regression (LR) and SVM classifiers, and for FG they use a centroid baseline and a regression model based on $k$-nearest neighbors alongside a decision tree regressor. 
Results averaged across five runs with random seeds for shuffling the data and initializing model parameters are presented. For CG, AlBERTo achieves best results, and LR performs the worst. SVM proves to be competitive for the task. In FG's case, AlBERTo achieves the best scores again. Interestingly, the decision tree performs competitively despite being a much more cost-efficient system. 

\paragraph{Portuguese.} \citet{pichel-campos-etal-2018-measuring} aims to compare six time periods of historical European Portuguese. They implement a perplexity-based language distance (PLD) measure with 7-gram models alongside a linear interpolation based smoothing technique. They conduct experiments on two levels: PLD with original spelling, and PLD with transcribed spelling. For the first instance, they compute PLD for each possible train-test pair. 
For the latter instance, they adjust the Diachronic Portuguese Corpus to have all periods share the same spelling. This is achieved by transliterating all historical periods into Latin scripts and then normalizing it with a generic orthography similar to phonological style. The resulting encoding of spelling normalization consists of 34 symbols, including 10 vowels and 24 consonants.

Overall, the results in both experiments observe a similar pattern. 
It is shown that language distances between different time periods are correlated with chronology. Moreover, there is not a huge divergence within the different periods investigated.  The longest difference between periods scores roughly 6.19 with original spelling and 5.92 with transcribed spelling, which is still lower than the distance between closely related languages, such as Spanish-Portuguese's score of 7.74. The results suggest that, at least for the case of Brasilian-Portuguese, the language has remained similar over time.

For the other study that works with Portuguese, \citet{zampieri-etal-2016-modeling} conduct experiments in three steps. They first have a preliminary session where they test a small sample with 87 documents from their corpus. They train SVM alongside Multinominal Naive Bayes (MNB) to predict which century a text belongs to, using both words and Part-of-Speech (POS) tags. 

Secondly, they start their main experiments where they use 1500 artificially generated documents, and use the SVM classifier to execute predictions. They observe 
a performance increase due to the implementation of POS tags or  words represented as uni-, bi-, and trigrams. 
Results show that POS trigrams yield the 90.7\% accuracy when tested with century classification of the presented documents. \citet{zampieri-etal-2016-modeling} note that this emphasises the existence of difference in structural properties in each time span by an important level; this means 
changes in structural properties take place at both the word level and beyond, and these changes can be captured through uni-, bi-, and trigrams.

Lastly, they conduct 
experiments across a smaller time span of 50 years. Their findings show that many time periods exhibit high similarity in grammatical structure. This presents a challenge for century classification of documents.
It is noted that POS tags perform the best with trigrams.

\paragraph{Swiss German.}
\citet{Jeszenszky_Stoeckle_Glaser_Weibel_2018} conceptionalize transitions between dialectal variant areas via logistic regression and intensity maps in an attempt to present spatial distribution of syntactic variants in Swiss. The results show gradual and sharp transitions between variants alongside distinct spatial patterns. Subdivision analyses further elucidated the characteristics of dominance zones and transition areas. Overall, the findings shed light on the spatial distribution and dynamics of linguistic features. A drawback of the methodology is that only 40\% of the variables in the SADS dataset \citep{GlaserBart+2015+81+108} can be modeled.
An important take-away is that the transition of dialectal variants is a highly complex phenomenon, which cannot be fully modeled by only taking the spatial dimension into account.

\citet{jeszenszky-etal-2019-spatio} use logistic regression on a global level to model the association of linguistic variation and age with 10-fold cross-validation. The AUC scores (area under the curve) reveal that for more than half of the variants considered, age is not a significant predictor. On a local level, they classify whether a specific linguistic variant is used at a survey site given the respondent age. The survey site is chosen from the $k$-nearest neighbors based on Euclidean distance, $k$ ranging from 5 to 50. 
They conclude that the significance of age as predictor variable is correlated with space: 
When a specific age group within a region is significant, the prediction of which dialect that region speaks is more accurate. However, the prediction becomes less accurate when a region associates with multiple age groups.
They attribute this finding to a sociolinguistic fact that lexicon in dialects is more prone to change with respect to speaker age than syntax.

\paragraph{German.}
\citet{dipper-waldenberger-2017-investigating} and \citet{WaldenbergerDipperLemke+2021+401+420} combine quantitative with an in-depth qualitative analysis. Both do not experiment with complex methods, but conduct a simple frequency-based, statistical analysis. 
\citet{dipper-waldenberger-2017-investigating} find quantitative proof for morphological, phonological, and graphemic phenomena by deriving replacement rules. 
They show insightful results  
into nuances of linguistic change across different regions and periods from a historical linguistic perspective: finding quantitative prove for theories like the High German consonant shift.
The second study by \citet{WaldenbergerDipperLemke+2021+401+420} employs slightly more elaborate statistical measures to quantify differences between texts and subcorpora. The results confirm the diatopic and diachronic variation: By analyzing Levenshtein mappings and computing similarity scores, the study demonstrated that texts from closely related dialects exhibited higher similarity scores compared to those from more distant regions. Overall, Upper German texts are found to be more similar to each other than Middle German texts.

\paragraph{English and French.}
\citet{montariol-allauzen-2021-measure} experiment with two kinds of embeddings, continuous bag-of-words (CBOW) and BERT \citep{devlin-etal-2019-bert}, to detect whether meaning changes of a word and its translation in English and French are consistent or divergent over time.
They show a trade-off between performance and efficiency: While BERT with k-means clustering achieves the best performance, the CBOW model with incremental training is computationally the most efficient and offers very competitive results.

Their findings are summarized as follows:
Semantic meanings drifting in the same direction across languages mainly occurs with words related to technology and society. On the other hand, meanings diverging in different directions implies that the meaning of a word might remain unchanged over time in one language, but drift in the other. This is mostly seen for words related to 
culture-specific concepts or controversial topics. 
It would be interesting to apply this approach not only to related languages, but to an actual dialect continuum to investigate whether these findings
are confirmed in closer related language varieties as well.

\section{Discussion}

Almost all languages in the world have distinct dialects varying by location that change quickly due to complex factors related to contact. Taking these two dimensions of language variation, diachronic and diatopic, into account can improve the diversity and representativeness of languages covered in this field
, and benefit the communities of non-standard language users. Our research shows that the intersection of diachronic and diatopic variation is an under-studied topic in dialect NLP. Although there are some 
approaches experimenting with diachronic word embeddings on a multilingual level \cite{montariol-allauzen-2021-measure}, there is currently a lack of state-of-the-art machine learning and NLP approaches.

This is a challenging topic to work with, considering 
its interdisciplinary nature combining historical linguistics, dialectology, machine learning and NLP. Perhaps this is a factor contributing to the status of deep learning based NLP methods having not yet been applied to studying language change in dialect continua.

\subsection{Open Challenges}
\label{sec:challenges}
\paragraph{Do language variants and dialects change over time?}
While \citet{pichel-campos-etal-2018-measuring} show that the difference in perplexity-based language distances between different time periods of European Portuguese is not substantial,  
\citet{zampieri-etal-2016-modeling} suggest that grammatical structure can be substantially different in some time periods of Portuguese; however, their study was conducted on artificial documents. This means that either perplexity-based language distance fails to capture the differences in grammatical structures of different time periods, or such differences are not present in the real-world historical Portuguese documents investigated. We leave this question to future work.  

\paragraph{Is the construction of dialect-related datasets reliable?}
The reliability of \citet{ramponi-casula-2023-diatopit} is also worth mentioning: They rely on the belief that the locals may write things that deviate from Standard Italian just because they speak it so, but they also rely on Twitter language identifiers to deduct whether a tweet is in Italian or not. This, of course, is a double-edged sword and may cut back on data reliability. If their assumption is correct, in extreme cases some societies whose language use deviate from the standard may remain completely under-represented and their twitters might be misclassified as Standard Italian. However, if it is incorrect (i.e., the language use of the locals follows the standard), the tweets written by the locals and those in standard Italian become indistinguishable.
Considering their access to speakers of regional Italian varieties (curators), 
\citet{koprivova-etal-2014-mapping} set a good example they could follow to ensure more varieties are correctly represented. However, one can argue that if someone was to use VPN for any reason, the coordinates would also be set for the entire time of use. Thus, Twitter APIs may not provide completely accurate data either,
though this may be minimal to consider in most cases.

\paragraph{Are speaker characteristics important?}
Additionally, although \citet{koprivova-etal-2014-mapping} show that tracking the number of generations present in a conversation is beneficial for building speaker-characters, \citet{jeszenszky-etal-2019-spatio} suggest that age is not a definitive for prediction. This means the usefulness of age information is quite task-dependent. An interesting follow-up work would be to incorporate other speaker characteristics, such as gender and education, into the analysis. 

\paragraph{Limited coverage of dialects.} 
There are numerous dialects spoken in the world. For instance, English alone has approximately 160 dialects \cite{aeni2021literature}. However, only a small number of dialects have been researched in the NLP and machine learning communities. Future work could establish a data center to manage and update world-existing dialect corpora. Indeed, many corpora are publicly available but are little explored.
For instance, the German regional newspaper corpus ZDL-Regionalkorpus \citep{NoldaBarbaresiGeyken2021} has not been used for diachronic analysis so far, despite its size of more than 11 B tokens
covering a time span 1993-2024 with regular updates 
which could enable use for data-intensive machine learning and word embedding approaches.

\paragraph{Ethical considerations in the collection of dialectal data.}
Although the data collection methods of \citet{koprivova-etal-2014-mapping} promise to provide near authentic results, no ethical issues are mentioned. As the conversations are recorded without the knowledge of the participants to ensure natural quality, it would not have been possible to get individual consent from the participants, although the person recording may have agreed otherwise. This, therefore, shows risk of privacy breach, and may not be an acceptable approach in a lot of data collection cases. Whether this would be acceptable if the speakers are informed after the data is collected may still be questionable to some people's discretion, however, this doesn't change the fact that despite being a breach, \citet{koprivova-etal-2014-mapping}'s approach does provide data as close to real-life situations as possible. This is of value in itself.

\section{Conclusion}
While there is a rising interest in modeling diachronic and diatopic variation in the NLP community, the intersection of both, i.e. language change in dialect continua, remains an under-studied topic. Even though findings from linguistics and sociolinguistics stress the importance of the diatopic dimension when modeling language change, the topic has not yet received as much attention in computational linguistics and not many methodological advancements have been made.
Our work has been a first step in closing this research gap, and we hope to give inspiration to future research.

\section*{Acknowledgements}
We thank the anonymous reviewers for their thoughtful feedback that greatly improved the texts. 
\bibliography{custom}

\begin{thebibliography}{41}
\providecommand{\natexlab}[1]{#1}

\bibitem[{Aeni et~al.(2021)Aeni, Octaberlina, Lubis et~al.}]{aeni2021literature}
Nur Aeni, Like~Raskova Octaberlina, Nenni Dwi~Aprianti Lubis, et~al. 2021.
\newblock A literature review of english language variation on sociolinguistics.

\bibitem[{Baldwin(2018)}]{baldwin-2018-language}
Timothy Baldwin. 2018.
\newblock \href {https://aclanthology.org/W18-3908} {Language and the shifting sands of domain, space and time (invited talk)}.
\newblock In \emph{Proceedings of the Fifth Workshop on {NLP} for Similar Languages, Varieties and Dialects ({V}ar{D}ial 2018)}, page~76, Santa Fe, New Mexico, USA. Association for Computational Linguistics.

\bibitem[{Beeching(2006)}]{beeching2006synchronic}
Kate Beeching. 2006.
\newblock Synchronic and diachronic variation: the how and why of sociolinguistic corpora.
\newblock In \emph{Corpus linguistics around the world}, pages 49--61. Brill.

\bibitem[{Boldsen and Paggio(2022)}]{boldsen-paggio-2022-letters}
Sidsel Boldsen and Patrizia Paggio. 2022.
\newblock \href {https://doi.org/10.18653/v1/2022.acl-long.463} {Letters from the past: Modeling historical sound change through diachronic character embeddings}.
\newblock In \emph{Proceedings of the 60th Annual Meeting of the Association for Computational Linguistics (Volume 1: Long Papers)}, pages 6713--6722, Dublin, Ireland. Association for Computational Linguistics.

\bibitem[{Cathcart and Wandl(2020)}]{cathcart-wandl-2020-search}
Chundra Cathcart and Florian Wandl. 2020.
\newblock \href {https://doi.org/10.18653/v1/2020.sigmorphon-1.28} {In search of isoglosses: continuous and discrete language embeddings in {S}lavic historical phonology}.
\newblock In \emph{Proceedings of the 17th SIGMORPHON Workshop on Computational Research in Phonetics, Phonology, and Morphology}, pages 233--244, Online. Association for Computational Linguistics.

\bibitem[{Chen et~al.(2024)Chen, Zhao, Breitbarth, Stoeckel, Mehler, and Eger}]{chen2024syntactic}
Yanran Chen, Wei Zhao, Anne Breitbarth, Manuel Stoeckel, Alexander Mehler, and Steffen Eger. 2024.
\newblock Syntactic language change in english and german: Metrics, parsers, and convergences.
\newblock \emph{arXiv preprint arXiv:2402.11549}.

\bibitem[{Conneau et~al.(2020)Conneau, Khandelwal, Goyal, Chaudhary, Wenzek, Guzm{\'a}n, Grave, Ott, Zettlemoyer, and Stoyanov}]{conneau-etal-2020-unsupervised}
Alexis Conneau, Kartikay Khandelwal, Naman Goyal, Vishrav Chaudhary, Guillaume Wenzek, Francisco Guzm{\'a}n, Edouard Grave, Myle Ott, Luke Zettlemoyer, and Veselin Stoyanov. 2020.
\newblock \href {https://doi.org/10.18653/v1/2020.acl-main.747} {Unsupervised cross-lingual representation learning at scale}.
\newblock In \emph{Proceedings of the 58th Annual Meeting of the Association for Computational Linguistics}, pages 8440--8451, Online. Association for Computational Linguistics.

\bibitem[{Devlin et~al.(2019)Devlin, Chang, Lee, and Toutanova}]{devlin-etal-2019-bert}
Jacob Devlin, Ming-Wei Chang, Kenton Lee, and Kristina Toutanova. 2019.
\newblock \href {https://doi.org/10.18653/v1/N19-1423} {{BERT}: Pre-training of deep bidirectional transformers for language understanding}.
\newblock In \emph{Proceedings of the 2019 Conference of the North {A}merican Chapter of the Association for Computational Linguistics: Human Language Technologies, Volume 1 (Long and Short Papers)}, pages 4171--4186, Minneapolis, Minnesota. Association for Computational Linguistics.

\bibitem[{Dipper and Schultz-Balluff(2013)}]{dipper2013anselm}
Stefanie Dipper and Simone Schultz-Balluff. 2013.
\newblock The anselm corpus: Methods and perspectives of a parallel aligned corpus.
\newblock In \emph{Proceedings of the workshop on computational historical linguistics at NODALIDA}, pages 27--42.

\bibitem[{Dipper and Waldenberger(2017)}]{dipper-waldenberger-2017-investigating}
Stefanie Dipper and Sandra Waldenberger. 2017.
\newblock \href {https://doi.org/10.18653/v1/W17-1204} {Investigating diatopic variation in a historical corpus}.
\newblock In \emph{Proceedings of the Fourth Workshop on {NLP} for Similar Languages, Varieties and Dialects ({V}ar{D}ial)}, pages 36--45, Valencia, Spain. Association for Computational Linguistics.

\bibitem[{Dunn and Wong(2022)}]{dunn-wong-2022-stability}
Jonathan Dunn and Sidney Wong. 2022.
\newblock \href {https://aclanthology.org/2022.coling-1.3} {Stability of syntactic dialect classification over space and time}.
\newblock In \emph{Proceedings of the 29th International Conference on Computational Linguistics}, pages 26--36, Gyeongju, Republic of Korea. International Committee on Computational Linguistics.

\bibitem[{Eisenstein et~al.(2014)Eisenstein, O'Connor, Smith, and Xing}]{10.1371/journal.pone.0113114}
Jacob Eisenstein, Brendan O'Connor, Noah~A. Smith, and Eric~P. Xing. 2014.
\newblock \href {https://doi.org/10.1371/journal.pone.0113114} {Diffusion of lexical change in social media}.
\newblock \emph{PLOS ONE}, 9(11):1--13.

\bibitem[{Faisal et~al.(2024)Faisal, Ahia, Srivastava, Ahuja, Chiang, Tsvetkov, and Anastasopoulos}]{faisal2024dialectbench}
Fahim Faisal, Orevaoghene Ahia, Aarohi Srivastava, Kabir Ahuja, David Chiang, Yulia Tsvetkov, and Antonios Anastasopoulos. 2024.
\newblock \href {https://arxiv.org/abs/2403.11009} {Dialectbench: A nlp benchmark for dialects, varieties, and closely-related languages}.
\newblock \emph{Preprint}, arXiv:2403.11009.

\bibitem[{Glaser and Bart(2015)}]{GlaserBart+2015+81+108}
Elvira Glaser and Gabriela Bart. 2015.
\newblock \href {https://doi.org/doi:10.1515/9783110363449-005} {\emph{4. Dialektsyntax des Schweizerdeutschen}}, pages 81--108.
\newblock De Gruyter, Berlin, München, Boston.

\bibitem[{He and Zhao(2024)}]{he2024sound}
Siqi He and Wei Zhao. 2024.
\newblock Exploring sound change over time: A review of computational and human perception.
\newblock In \emph{Proceedings of the 5th Workshop on Computational Approaches to Historical Language Change}, Bangkok. Association for Computational Linguistics.

\bibitem[{Jeszenszky et~al.(2019)Jeszenszky, Siriaraya, Stoeckle, and Jatowt}]{jeszenszky-etal-2019-spatio}
P{\'e}ter Jeszenszky, Panote Siriaraya, Philipp Stoeckle, and Adam Jatowt. 2019.
\newblock \href {https://doi.org/10.18653/v1/W19-4723} {Spatio-temporal prediction of dialectal variant usage}.
\newblock In \emph{Proceedings of the 1st International Workshop on Computational Approaches to Historical Language Change}, pages 186--195, Florence, Italy. Association for Computational Linguistics.

\bibitem[{Jeszenszky et~al.(2018)Jeszenszky, Stoeckle, Glaser, and Weibel}]{Jeszenszky_Stoeckle_Glaser_Weibel_2018}
Péter Jeszenszky, Philipp Stoeckle, Elvira Glaser, and Robert Weibel. 2018.
\newblock \href {https://doi.org/10.1017/jlg.2019.1} {A gradient perspective on modeling interdialectal transitions}.
\newblock \emph{Journal of Linguistic Geography}, 6(2):78–99.

\bibitem[{Joshi et~al.(2024)Joshi, Dabre, Kanojia, Li, Zhan, Haffari, and Dippold}]{joshi2024natural}
Aditya Joshi, Raj Dabre, Diptesh Kanojia, Zhuang Li, Haolan Zhan, Gholamreza Haffari, and Doris Dippold. 2024.
\newblock \href {https://arxiv.org/abs/2401.05632} {Natural language processing for dialects of a language: A survey}.
\newblock \emph{Preprint}, arXiv:2401.05632.

\bibitem[{Komrskova et~al.(2017)Komrskova, Kopřivova, Lukeš, Poukarová, and Goláňová}]{komrskova}
Zuzana Komrskova, Marie Kopřivova, David Lukeš, Petra Poukarová, and Hana Goláňová. 2017.
\newblock \href {https://doi.org/10.1515/jazcas-2017-0031} {New spoken corpora of czech: Ortofon and dialekt}.
\newblock \emph{Journal of Linguistics/Jazykovedný casopis}, 68.

\bibitem[{Kop{\v{r}}ivov{\'a} et~al.(2014)Kop{\v{r}}ivov{\'a}, Gol{\'a}{\v{n}}ov{\'a}, Klime{\v{s}}ov{\'a}, and Luke{\v{s}}}]{koprivova-etal-2014-mapping}
Marie Kop{\v{r}}ivov{\'a}, Hana Gol{\'a}{\v{n}}ov{\'a}, Petra Klime{\v{s}}ov{\'a}, and David Luke{\v{s}}. 2014.
\newblock \href {http://www.lrec-conf.org/proceedings/lrec2014/pdf/252_Paper.pdf} {Mapping diatopic and diachronic variation in spoken {C}zech: The {ORTOFON} and {DIALEKT} corpora}.
\newblock In \emph{Proceedings of the Ninth International Conference on Language Resources and Evaluation ({LREC}'14)}, pages 376--382, Reykjavik, Iceland. European Language Resources Association (ELRA).

\bibitem[{Kutuzov et~al.(2018)Kutuzov, {\O}vrelid, Szymanski, and Velldal}]{kutuzov-etal-2018-diachronic}
Andrey Kutuzov, Lilja {\O}vrelid, Terrence Szymanski, and Erik Velldal. 2018.
\newblock \href {https://aclanthology.org/C18-1117} {Diachronic word embeddings and semantic shifts: a survey}.
\newblock In \emph{Proceedings of the 27th International Conference on Computational Linguistics}, pages 1384--1397, Santa Fe, New Mexico, USA. Association for Computational Linguistics.

\bibitem[{Ma et~al.(2024{\natexlab{a}})Ma, Schlechtweg, and Zhao}]{ma2024presence}
Xianghe Ma, Dominik Schlechtweg, and Wei Zhao. 2024{\natexlab{a}}.
\newblock Presence or absence: Are unknown word usages in dictionaries?
\newblock In \emph{Proceedings of the 5th Workshop on Computational Approaches to Historical Language Change}, Bangkok. Association for Computational Linguistics.

\bibitem[{Ma et~al.(2024{\natexlab{b}})Ma, Strube, and Zhao}]{ma-etal-2024-graph}
Xianghe Ma, Michael Strube, and Wei Zhao. 2024{\natexlab{b}}.
\newblock \href {https://aclanthology.org/2024.eacl-long.93} {Graph-based clustering for detecting semantic change across time and languages}.
\newblock In \emph{Proceedings of the 18th Conference of the European Chapter of the Association for Computational Linguistics (Volume 1: Long Papers)}, pages 1542--1561, St. Julian{'}s, Malta. Association for Computational Linguistics.

\bibitem[{Montanelli and Periti(2023)}]{montanelli2023survey}
Stefano Montanelli and Francesco Periti. 2023.
\newblock \href {https://arxiv.org/abs/2304.01666} {A survey on contextualised semantic shift detection}.
\newblock \emph{Preprint}, arXiv:2304.01666.

\bibitem[{Montariol and Allauzen(2021)}]{montariol-allauzen-2021-measure}
Syrielle Montariol and Alexandre Allauzen. 2021.
\newblock \href {https://doi.org/10.18653/v1/2021.acl-long.100} {Measure and evaluation of semantic divergence across two languages}.
\newblock In \emph{Proceedings of the 59th Annual Meeting of the Association for Computational Linguistics and the 11th International Joint Conference on Natural Language Processing (Volume 1: Long Papers)}, pages 1247--1258, Online. Association for Computational Linguistics.

\bibitem[{Nolda et~al.(2021)Nolda, Barbaresi, and Geyken}]{NoldaBarbaresiGeyken2021}
Andreas Nolda, Adrien Barbaresi, and Alexander Geyken. 2021.
\newblock \href {https://doi.org/10.1515/9783110731514-018} {{Das ZDL-Regionalkorpus: Ein Korpus f{\"u}r die lexikografische Beschreibung der diatopischen Variation im Standarddeutschen}}.
\newblock Deutsch in Europa. Sprachpolitisch, grammatisch, methodisch, pages 317 -- 321. de Gruyter, Berlin [u.a.].

\bibitem[{Nolda et~al.(2023)Nolda, Barbaresi, and Geyken}]{NoldaBarbaresiGeyken2023}
Andreas Nolda, Adrien Barbaresi, and Alexander Geyken. 2023.
\newblock \href {https://doi.org/10.1515/9783111085708-003} {{Korpora f{\"u}r die lexikographische Beschreibung diatopischer Variation in der deutschen Standardsprache. Das ZDL-Regionalkorpus und das Webmonitor-Korpus}}.
\newblock Korpora in der germanistischen Sprachwissenschaft. M{\"u}ndlich, schriftlich, multimedial, pages 29 -- 52. de Gruyter, Berlin/Boston.

\bibitem[{Parisi et~al.(2020)Parisi, Francia, and Magnani}]{musixmatch-2020-umberto}
Loreto Parisi, Simone Francia, and Paolo Magnani. 2020.
\newblock Umberto: an italian language model trained with whole word masking.
\newblock \url{https://github.com/musixmatchresearch/umberto}.

\bibitem[{Petran et~al.(2016)Petran, Bollmann, Dipper, and Klein}]{Petran_Bollmann_Dipper_Klein_2016}
Florian Petran, Marcel Bollmann, Stefanie Dipper, and Thomas Klein. 2016.
\newblock \href {https://doi.org/10.21248/jlcl.31.2016.208} {Rem: A reference corpus of middle high german -- corpus compilation, annotation, and access}.
\newblock \emph{Journal for Language Technology and Computational Linguistics}, 31(2):1–15.

\bibitem[{Pichel~Campos et~al.(2018)Pichel~Campos, Gamallo, and Alegria}]{pichel-campos-etal-2018-measuring}
Jose~Ramom Pichel~Campos, Pablo Gamallo, and I{\~n}aki Alegria. 2018.
\newblock \href {https://aclanthology.org/W18-3916} {Measuring language distance among historical varieties using perplexity. application to {E}uropean {P}ortuguese.}
\newblock In \emph{Proceedings of the Fifth Workshop on {NLP} for Similar Languages, Varieties and Dialects ({V}ar{D}ial 2018)}, pages 145--155, Santa Fe, New Mexico, USA. Association for Computational Linguistics.

\bibitem[{Polignano et~al.(2019)Polignano, Basile, De~Gemmis, Semeraro, Basile et~al.}]{polignano2019alberto}
Marco Polignano, Pierpaolo Basile, Marco De~Gemmis, Giovanni Semeraro, Valerio Basile, et~al. 2019.
\newblock Alberto: Italian bert language understanding model for nlp challenging tasks based on tweets.
\newblock In \emph{CEUR workshop proceedings}, volume 2481, pages 1--6. CEUR.

\bibitem[{Ramponi and Casula(2023)}]{ramponi-casula-2023-diatopit}
Alan Ramponi and Camilla Casula. 2023.
\newblock \href {https://doi.org/10.18653/v1/2023.vardial-1.19} {{D}iatop{I}t: A corpus of social media posts for the study of diatopic language variation in {I}taly}.
\newblock In \emph{Tenth Workshop on NLP for Similar Languages, Varieties and Dialects (VarDial 2023)}, pages 187--199, Dubrovnik, Croatia. Association for Computational Linguistics.

\bibitem[{Sloetjes and Wittenburg(2008)}]{sloetjes2008annotation}
Han Sloetjes and Peter Wittenburg. 2008.
\newblock Annotation by category-elan and iso dcr.
\newblock In \emph{6th international Conference on Language Resources and Evaluation (LREC 2008)}.

\bibitem[{S{\o}gaard et~al.(2021)S{\o}gaard, Ebert, Bastings, and Filippova}]{sogaard-etal-2021-need}
Anders S{\o}gaard, Sebastian Ebert, Jasmijn Bastings, and Katja Filippova. 2021.
\newblock \href {https://doi.org/10.18653/v1/2021.eacl-main.156} {We need to talk about random splits}.
\newblock In \emph{Proceedings of the 16th Conference of the European Chapter of the Association for Computational Linguistics: Main Volume}, pages 1823--1832, Online. Association for Computational Linguistics.

\bibitem[{Sukhareva and Chiarcos(2014)}]{sukhareva-chiarcos-2014-diachronic}
Maria Sukhareva and Christian Chiarcos. 2014.
\newblock \href {https://doi.org/10.3115/v1/W14-5302} {Diachronic proximity vs. data sparsity in cross-lingual parser projection. a case study on {G}ermanic}.
\newblock In \emph{Proceedings of the First Workshop on Applying {NLP} Tools to Similar Languages, Varieties and Dialects}, pages 11--20, Dublin, Ireland. Association for Computational Linguistics and Dublin City University.

\bibitem[{Vidal-Gor{\`e}ne et~al.(2020)Vidal-Gor{\`e}ne, Khurshudyan, and Donab{\'e}dian-Demopoulos}]{vidal-gorene-etal-2020-recycling}
Chahan Vidal-Gor{\`e}ne, Victoria Khurshudyan, and Ana{\"\i}d Donab{\'e}dian-Demopoulos. 2020.
\newblock \href {https://aclanthology.org/2020.vardial-1.9} {Recycling and comparing morphological annotation models for {A}rmenian diachronic-variational corpus processing}.
\newblock In \emph{Proceedings of the 7th Workshop on NLP for Similar Languages, Varieties and Dialects}, pages 90--101, Barcelona, Spain (Online). International Committee on Computational Linguistics (ICCL).

\bibitem[{Waldenberger et~al.(2021)Waldenberger, Dipper, and Lemke}]{WaldenbergerDipperLemke+2021+401+420}
Sandra Waldenberger, Stefanie Dipper, and Ilka Lemke. 2021.
\newblock \href {https://doi.org/doi:10.1515/zfs-2021-2037} {Towards a broad-coverage graphemic analysis of large historical corpora}.
\newblock \emph{Zeitschrift für Sprachwissenschaft}, 40(3):401--420.

\bibitem[{Wolfram and Schilling-Estes(2017)}]{wolfram_schillingch24}
Walt Wolfram and Natalie Schilling-Estes. 2017.
\newblock \href {https://doi.org/10.1002/9781405166201.ch24} {\emph{Dialectology and Linguistic Diffusion}}, chapter~24.
\newblock John Wiley \& Sons, Ltd.

\bibitem[{Zampieri and Becker(2013)}]{zampieri2013colonia}
Marcos Zampieri and Martin Becker. 2013.
\newblock Colonia: Corpus of historical portuguese.
\newblock \emph{ZSM Studien}, 5:69--76.

\bibitem[{Zampieri et~al.(2016)Zampieri, Malmasi, and Dras}]{zampieri-etal-2016-modeling}
Marcos Zampieri, Shervin Malmasi, and Mark Dras. 2016.
\newblock \href {https://aclanthology.org/L16-1647} {Modeling language change in historical corpora: The case of {P}ortuguese}.
\newblock In \emph{Proceedings of the Tenth International Conference on Language Resources and Evaluation ({LREC}'16)}, pages 4098--4104, Portoro{\v{z}}, Slovenia. European Language Resources Association (ELRA).

\bibitem[{Zampieri et~al.(2020)Zampieri, Nakov, and Scherrer}]{zampieri2020natural}
Marcos Zampieri, Preslav Nakov, and Yves Scherrer. 2020.
\newblock Natural language processing for similar languages, varieties, and dialects: A survey.
\newblock \emph{Natural Language Engineering}, 26(6):595--612.

\end{thebibliography}

\end{document}